\documentclass[runningheads]{llncs}
\usepackage{graphicx}
\usepackage{comment}
\usepackage{amsmath,amssymb}
\usepackage{color}
\usepackage{url}
\usepackage{hyperref}
\usepackage{caption}
\usepackage{subcaption}
\usepackage{arydshln}
\newif\ifreview
\reviewfalse
\ifreview
\usepackage{lineno}

\linenumbers
\fi

\begin{document}
\def\SubNumber{63}

\def\GCPRTrack{Review copy}
\title{FourierMask: Instance Segmentation using Fourier Mapping in Implicit Neural Networks}
%
%

%

\ifreview
\else
	\titlerunning{FourierMask}

	\author{Hamd ul Moqeet Riaz \and
		Nuri Benbarka \and
		Timon Hoefer
		\and Andreas Zell }
	\authorrunning{H. Riaz et al.}
	%
	\institute{Department of Computer Science (WSI), University of Tuebingen, Germany \\
		\url{https://uni-tuebingen.de/de/118829} 
		\\
		\email{\{hamd.riaz, nuri.benbarka, timon.hoefer, andreas.zell\}@uni-tuebingen.de}}
\fi

\maketitle              
\begin{abstract}
	We present FourierMask, which employs Fourier series combined with implicit neural representations to generate instance segmentation masks. We apply a Fourier mapping (FM) to the coordinate locations and utilize the mapped features as inputs to an implicit representation (coordinate based multi layer perceptron (MLP)). FourierMask learns to predict the coefficients of the FM for a particular instance, and therefore adapts the FM to a specific object. This allows FourierMask to be generalized to predict instance segmentation masks from natural images. Since implicit functions are continuous in the domain of input coordinates, we illustrate that by sub-sampling the input pixel coordinates, we can generate higher resolution masks during inference. Furthermore, we train a renderer MLP (\textit{FourierRend}) on the uncertain predictions of FourierMask and illustrate that it significantly improves the quality of the masks. FourierMask shows competitive results on the MS COCO dataset compared to the baseline Mask R-CNN at the same output resolution and surpasses it on higher resolution. 
	
\end{abstract}
\section{Introduction}
In the past decade, we have witnessed a shift from classical approaches towards deep learning methods for a variety of real world tasks. Due to an ample amount of real and synthetic datasets and high computation power, it has been possible to reliably use these 'black box' models on highly complex and critical use cases. For the field of autonomous driving, perceiving and understanding the environment is crucial. Instance segmentation is one of those tasks which allows autonomous systems to semantically separate different regions in their percepts (images) and at the same time separate objects from each other. 

In recent years, the majority of the methods have employed CNNs for the task of instance segmentation. There are methods that generate instance segmentation masks by classifying each pixel of a region of interest as either foreground or background e.g Mask R-CNN \cite{he2017mask}. These methods generally show the best performance but suffer from high computation needs and slow speed. There are methods that predict the contour points around the boundary of the object \cite{yang2019dense} \cite{xie2020polarmask}. 
Although being faster, these methods fail to match the performance of pixel-wise classification methods. Alternatively, there are methods that try to encode the mask contours in a compressed representation \cite{Xu_2019_ESE} \cite{riaz2020fouriernet}. These mask representations are compact and meaningful but lack the superior capabilities of pixel-wise methods.
\begin{figure}[t]
	\vspace{-0.2cm}	
	\centering
	\begin{subfigure}[b]{0.3\textwidth}
		\centering
		\includegraphics[width=\textwidth]{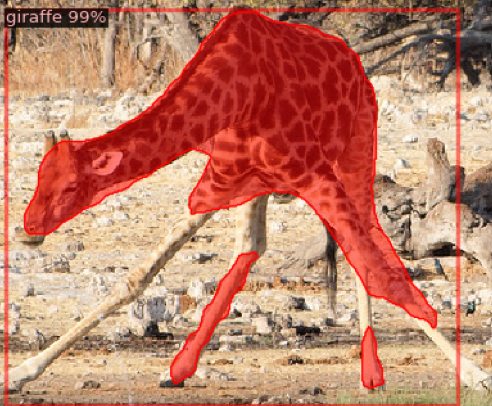}
		\caption{Mask R-CNN}
		\label{fig:maskrcnn}
	\end{subfigure}
	\begin{subfigure}[b]{0.3\textwidth}
		\centering
		\includegraphics[width=\textwidth]{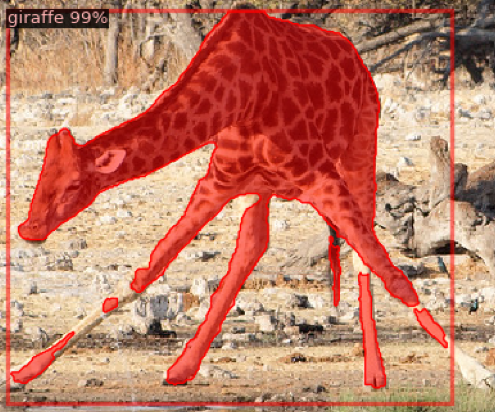}
		\caption{PointRend}
		\label{fig:pointrend}
	\end{subfigure}
	\begin{subfigure}[b]{0.31\textwidth}
		\centering
		\includegraphics[width=\textwidth]{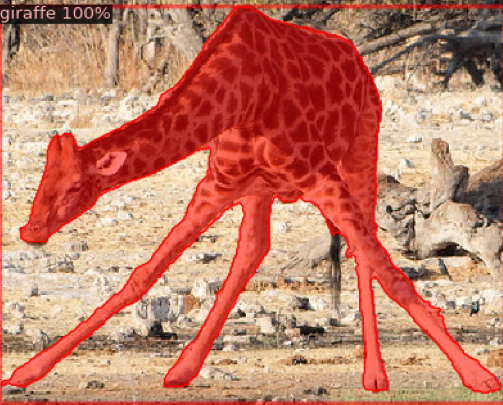}
		\caption{FourierRend}
		\label{fig:fouriermask}
	\end{subfigure}
	\caption{Comparison between Mask R-CNN \cite{he2017mask}, PointRend \cite{Kirillov_2020_CVPR} and FourierMask.}
	\label{fig:vis_compare}
	\vspace{-0.3cm}	
\end{figure}
Our approach employs a pixel-wise mask representation and additionally we generate the mask using a compact Fourier representation. In the case of segmentation masks, the Fourier series' low-frequency components hold the general shape and high-frequency components hold the edges of the mask. Therefore, our representation is meaningful and can be compressed according to the use case.

FourierMask is an \textit{implicit neural representation}. For the image regression task, \textit{implicit representations} learn to predict the RGB values of a particular image given a pixel coordinate. Tancik et al. \cite{tancik2020fourier} showed that using a Fourier Mapping of coordinates instead of actual coordinates locations as inputs, allows the implicit networks to learn high-frequency details in images and 3D scenes. Our work draws inspiration from this work and applies their findings to the task of generating masks for instance segmentation. Implicit representations have the advantage of learning and reconstructing fine details, which traditional representations cannot do as effectively in such compact models. As implicit functions are continuous in the domain of input coordinates, we can sub-sample the pixel coordinates to generate higher resolution masks during inference. However, these representations are inherently trained on a single image and have not yet been adopted to a general task of instance segmentation on natural images. 
Our contributions are as follows:
\begin{enumerate}
	\item We developed FourierMask, which can replace any mask predictor that uses a region of interest (ROI) to predict a binary mask. It is fully differentiable and end-to-end trainable. 
	\item We show that implicit representations can be applied to the task of instance segmentation. We achieve this by learning the coefficients of the Fourier mapping of a particular object.
	\item As implicit functions are continuous in the domain of input coordinates, we show that we can sub-sample the pixel coordinates to generate higher resolution masks during inference. These higher resolution masks are smoother and improve the performance on MS COCO. 
	\item We verify and illustrate that the rendering strategy from PointRend \cite{Kirillov_2020_CVPR} brings significant qualitative gains for FourierMask. Our renderer MLP \textit{FourierRend} improves the mask boundary of FourierMask significantly.
\end{enumerate}
\section{Related work}
\label{subsec:tsm}
In \textit{two-stage} instance segmentation, the network first detects (proposes) the objects and then predicts a segmentation mask from the detected region. The baseline method for many two stage methods is Mask R-CNN \cite{he2017mask}. Mask R-CNN added a mask branch to Faster R-CNN \cite{ren2015faster}, which generated a binary mask that separated the foreground and background pixels in a region of interest. 
Mask Scoring R-CNN \cite{huang2019mask} had a network block to learn the quality of the predicted instance masks and regressed the mask IoU. 
ShapeMask \cite{kuo2019shapemask} estimated the shape from bounding box detections using shape priors and refined it into a mask by learning instance embeddings. 
CenterMask \cite{lee2020centermask} added spatial attention-guided mask on top FCOS \cite{tian2019fcos} object detector, which helped to focus on important pixels and diminished noise.  PointRend \cite{Kirillov_2020_CVPR} tackled instance segmentation as rendering problem. By sampling unsure points from the feature map and its fine grained features from higher resolution feature map, it was able to predict really crisp object boundaries using a fully connected MLP. 
Rather than employing binary grid representation of masks, PolyTransform \cite{liang2020polytransform} used a polygon representation. These methods accomplish state-of-the-art accuracy, but they are generally slower than one stage methods.

\textit{One stage} instance segmentation methods predict the instance masks in a single shot, without using any proposed regions/bounding boxes as an intermediate step. YOLACT \cite{bolya2019yolact} linearly combined prototype masks and mask coefficients for each instance, to predict masks in real-time speeds. Likewise, Embedmask \cite{ying2019embedmask} employed embedding modules for pixels and proposals. ExtremeNet \cite{zhou2019bottom} predicted the contour (octagon) around an object using keypoints of the object. Similary, Polarmask \cite{xie2020polarmask} used the polar representation to predict a contour from a center point (centerness from FCOS \cite{tian2019fcos}). 
Dense RepPoints \cite{yang2019dense} used a large set of points to represent the boundary of objects. FourierNet \cite{riaz2020fouriernet} employed inverse fast Fourier transform (IFFT) to generate a contour around an object represented by polar coordinates. The network learned the coefficients of the Fourier series to predict those contours. 

\textit{Implicit representations} learn to encode a signal as a continuous function. Mescheder et al. \cite{mescheder2019occupancy} proposed occupancy networks, which implicitly represented the 3D surface as the continuous decision boundary of a deep neural network classifier. Mildenhall et al. \cite{mildenhall2020nerf} used implicit neural networks to learn 3D scenes and synthesize novel-views. Tancik et al. \cite{tancik2020fourier} showed that mapping input coordinates using a Fourier feature mapping, allows MLPs to learn high frequency function in low dimensional domains.  Sirens \cite{sitzmann2019siren} showed that using periodic activation functions (such as sine) in implicit representations, lets the MLPs learn the information of natural signals and their derivatives better than when using other activation functions such as ReLU. 
\section{Method}
\begin{figure*}[t]
	\vspace{-0.2cm}	
	\begin{center}
		\includegraphics[width=0.8\linewidth]{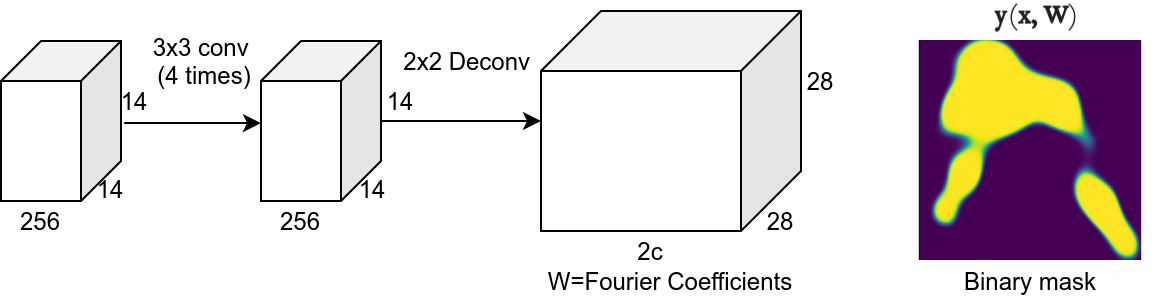}
	\end{center}
	\caption{FourierMask head architecture for a ROI Align size of 14x14. The network predicts Fourier coefficients $\textbf{W}$ for each location in the feature map.}
	\label{fig:FourierMaskHead}
	\vspace{-0.2cm}	
\end{figure*}
\subsection{FourierMask head architecture}
This section explains the network architecture of FourierMask.
We employ Mask R-CNN \cite{he2017mask} as our baseline model. We use a ResNet \cite{He_2016_CVPR} backbone pre-trained on ImageNet dataset \cite{deng2009imagenet}, with a feature pyramid network (FPN) \cite{Lin_2017_CVPR} architecture. Following from Mask R-CNN, we use a small region proposal network (RPN), which generates $k$ proposal candidates from all feature levels from FPN. To generate fixed size feature maps from these proposal candidates, we use a ROI Align \cite{he2017mask} operation. This produces a $(k, d, m, m)$ sized feature map for the mask head, where $m$ is the fixed spatial size after ROI Align operation and $d$ is the number of channels. The structure of the head is shown in the figure \ref{fig:FourierMaskHead}. We apply four convolutions consecutively, each with a kernel size $3\times 3$ and a stride of 1. Then we apply a transposed convolution layer with $2c$ number of filters, which generates a spatial volume of size $2c \times 2m\times 2m$. We call this feature volume $\textbf{W}$, which holds $2c$ Fourier coefficients for each spatial location. 
\subsection{Fourier Features}
\label{ssec:fouriermapping}
In this section, we explain how to obtain the Fourier features from the coefficients $\textbf{W}$. First, we define the Fourier mapping as 
\begin{equation}
\label{eq:FourierMapping}
\gamma(\textbf{x}) = [cos(2\pi \textbf{x}\cdot \textbf{B}), sin(2\pi \textbf{x}\cdot \textbf{B})].
\end{equation}
Here $\textbf{x} \in \mathbb{R}^{p \times 2}$ are the pixel coordinates $(i,j)$ normalized to a value in range $[0,1]$ and $p$ are the total number of pixels in the image. Since sine and cosine have a period of $2\pi$, by normalizing the pixel coordinate $\textbf{x}$ to a range $[0,1]$, we ensure one complete image lies in period of $2\pi$. Images are not periodic signals, but since they are bounded by an image resolution, we can safely apply our method to predict 2D binary masks. 
$\textbf{B} \in \mathbb{Z}^{2\times c}$ is the integer lattice matrix which holds the possible combinations of harmonic frequency integers of Fourier series for both dimensions in the image.
$\textbf{B}$ contains the elements of the set \textit{S}, defined by
\begin{align}\label{def:B}
\begin{split}
S=\{0,\dots,f\} \times \{-f,\dots,f\}
\setminus\{ 0\} \times \{-f,\dots,-1\},
\end{split}
\end{align}
where $f$ is the total number of frequencies. \textbf{B} has non-negative integers in its first row and integers in its second row, except that the second row wont have negative integers for a zero in the first row.  
This way of defining \textbf{B} is motivated by the 2D Fourier representation. If we do not limit \textbf{B} by the total number of frequencies $f$, we would obtain the original 'sine plus cosine' form of the Fourier series. This property is shown in the supplementary material. However, we have to limit $f$ to a number which fits to our memory constraints and speed requirements. For the case of images (2D input), the possible permutations  $c$ can be calculated by:
\begin{equation}
\label{eq:NumofCoe}
c = (f + 1)(2f + 1)  - f
\end{equation}
Fourier Features are generated as follows: 
\begin{align}
\label{eq:FourierFeatures}
\textbf{FF(x, W)} = \gamma(\textbf{x})\circ \textbf{W},
\end{align}
where $\circ$ is the element-wise product, $\textbf{W}\in \mathbb{R}^{p \times 2c}$ is the weight matrix predicted by the FourierMask. Note that we flatten the spatial dimension of the prediction beforehand ($p=2m\times2m$).  Let $\textbf{\textit{ff}}_i$ be the $i_{th}$ column of $\textbf{FF}$; then the binary mask $\textbf{y}$ is defined as
\begin{align}
\label{eq:FF_output}
\textbf{y}(\textbf{x}, \textbf{W}) = \phi ( \sum_{i=1}^{2c} \textbf{\textit{ff}}_i) .
\end{align}
Here $\phi$ is the sigmoid activation function, which we use to bound the output between 0 and 1. Note that $\textbf{y}(\textbf{x}, \textbf{W})$ can be interpreted as an implicit representation with a single perceptron because it is a linear combination of Fourier features followed by a non linear activation function.
\subsection{Fourier Features based MLP}
\label{ssec:FFMLP}
As shown by \cite{sitzmann2019siren,tancik2020fourier}, implicit representations can learn to generate shapes, images etc. from input coordinates very effectively. Following the work from \cite{tancik2020fourier}, we saw that Fourier mapping of input coordinates lets the MLP learn higher frequencies and consequently generate images with finer detail compared to MLPs without Fourier mapping. Furthermore, \cite{sitzmann2019siren} showed that using periodic activation functions works better compared to ReLU in implicit neural networks. We employ an MLP with sine activation functions, in which Fourier features (FF) are the input and mask $\textbf{y}^\prime$ is the output. We have 3 hidden layers (Siren layers), each with 256 neurons. The MLP has a single output neuron, on which we apply a sigmoid function to bound it between 0 and 1. The Fourier features (FF) are generated by the equation \eqref{eq:FourierFeatures} and they are parameterized by coefficients $\textbf{W}$ learned by the network and therefore adapted for a specific input ROI. Coordinate based MLPs encode the information of one particular image or shape, but by parameterizing them with learned Fourier coefficients $\textbf{W}$ of each object, we can generalize them to generate a binary mask of any object. 
\subsection{MLP as a renderer - FourierRend}
\label{ssec:mlp_renderer}
\begin{figure}[b]
	\vspace{-0.2cm}	
	\begin{center}
		\includegraphics[width=0.75\linewidth]{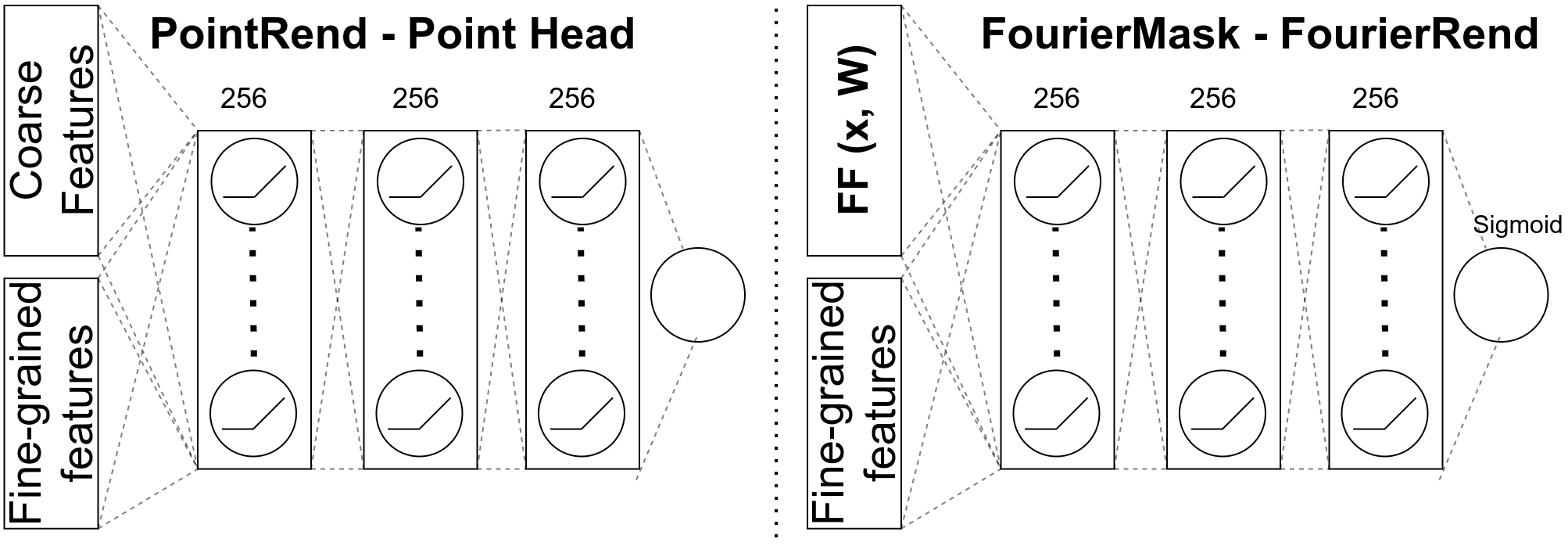}
	\end{center}
	\caption{Difference between point head from PointRend and FourierRend.}
	\label{fig:fourirerend}	
	\vspace{-0.3cm}	
\end{figure}
For generating boundary-aligned masks, we used a renderer MLP (FourierRend) which specialized only on the uncertain regions of the mask predicted by equation \eqref{eq:FF_output}. We adopted the rendering strategy from PointRend \cite{Kirillov_2020_CVPR} and made the following modification in the point head (figure \ref{fig:fourirerend}). Rather than sampling coarse mask features in the mask head, we sample the Fourier features (\textbf{FF}) from equation \eqref{eq:FourierFeatures} at uncertain mask prediction coordinates (the locations where the predictions are near 0.5). Fourier features \textbf{FF(x, W)} makes FourierRend an implicit MLP since its input is a function of input coordinates \textbf{x} and therefore we can take leverage from its implicit nature as discussed before. We concatenate these Fourier features and fine-grained features (from the 'p2' level FPN feature map). We replace the mask predictions from equation \eqref{eq:FF_output} (coarse predictions), with the fine-detailed predictions from FourierRend. Consequently, we replace uncertain predictions at the boundary, with more accurate predictions of the renderer, resulting in crisp and boundary-aligned masks. 
\subsection{Training and loss function}
We concatenate the output $\textbf{y}$ from equation \eqref{eq:FF_output} and $\textbf{y}^\prime$ from the MLP and train both masks in parallel. By training the output $\textbf{y}$, we learn the coefficients $\textbf{W}$ of a Fourier series in their true sense. We need these coefficients because we assume that the input for the MLP are Fourier features.
We use IoU loss for training the binary masks defined as:
\begin{equation}
\textbf{IoU loss} = \frac{\sum_{i=0}^{N} \min(y_{p_i}, y_{t_i})}{\sum_{j=0}^{N} \max(y_{p_j}, y_{t_j})}
\end{equation}
$y_{p_i}$ is the predicted value of the pixel $i$, $y_{t_i}$ is the ground truth value of the pixel $i$ and $N$ is the total number of pixels in the predicted mask. 

\section{Experiments}
For all our experiments we employ a Resnet 50 backbone with feature pyramid network pre-trained on ImageNet \cite{deng2009imagenet} unless otherwise stated. We use the Mask R-CNN default settings from detectron2 \cite{wu2019detectron2}. We train on the MS COCO \cite{lin2014microsoft} training set and show the results on its validation set. We predict class agnostic masks, i.e. rather than predicting a mask for each class in MS COCO, we predict only one mask per ROI. For the baseline, we trained a Mask R-CNN and PointRend \cite{Kirillov_2020_CVPR} with class agnostic masks.
\subsection{Spectrum Analysis MS COCO}
\label{ssec:spectrum}

\begin{figure}[t]
	\centering
	\vspace{-0.2cm}	
	\begin{minipage}[b]{0.45\textwidth}
		\centering
		\includegraphics[width=0.99\linewidth]{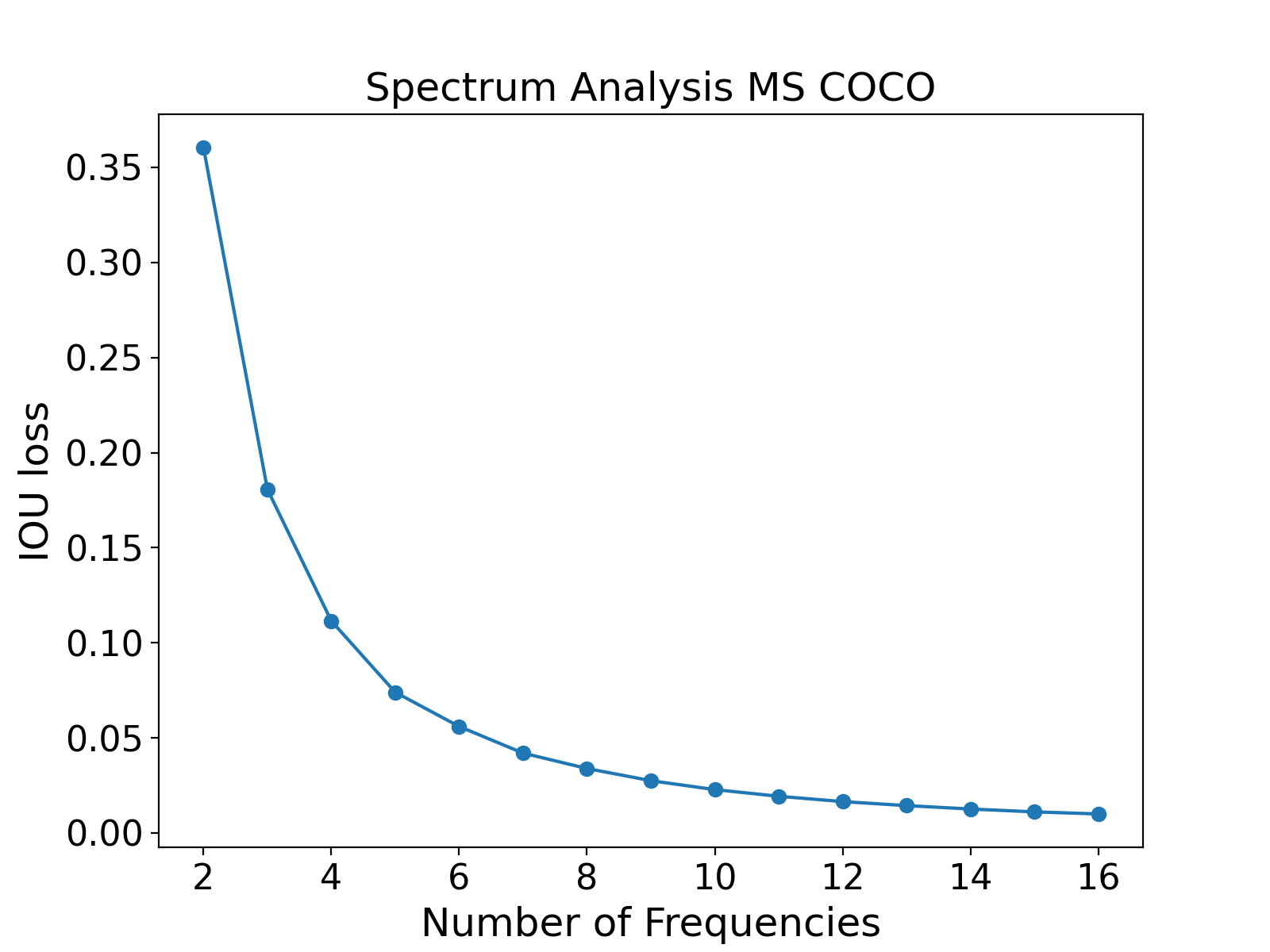}
		\caption{Spectrum test on COCO dataset.}
		\label{fig:Spectrum}
	\end{minipage}
\hspace{0.1cm}
	\begin{minipage}[b]{0.45\textwidth}
		\centering
		\begin{subfigure}[b]{0.25\textwidth}
			\centering
			\includegraphics[width=\textwidth]{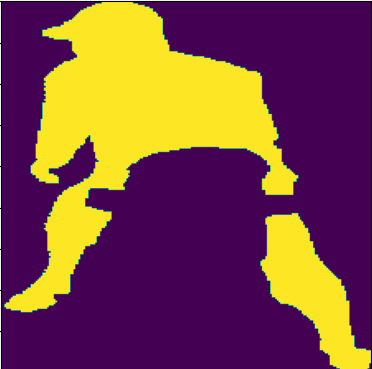}
			\caption{GT}
			\label{fig:spectrum_gt}
		\end{subfigure}
		\begin{subfigure}[b]{0.25\textwidth}
			\centering
			\includegraphics[width=\textwidth]{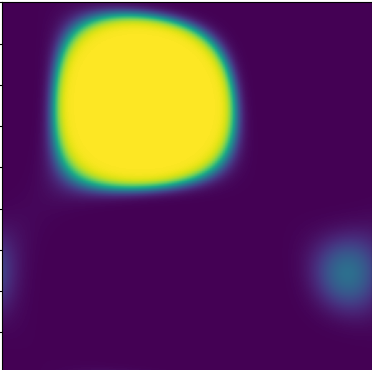}
			\caption{2}
			\label{fig:spectrum_2}
		\end{subfigure}
		\begin{subfigure}[b]{0.25\textwidth}
			\centering
			\includegraphics[width=\textwidth]{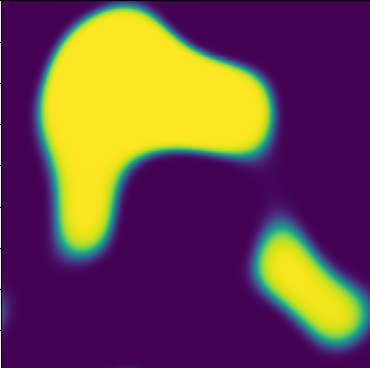}
			\caption{3}
			\label{fig:spectrum_3}
		\end{subfigure}
		\begin{subfigure}[b]{0.25\textwidth}
			\centering
			\includegraphics[width=\textwidth]{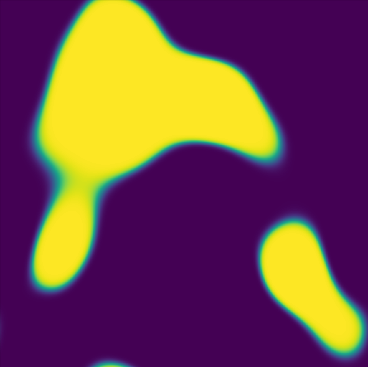}
			\caption{4}
			\label{fig:spectrum_4}
		\end{subfigure}
		\begin{subfigure}[b]{0.25\textwidth}
			\centering
			\includegraphics[width=\textwidth]{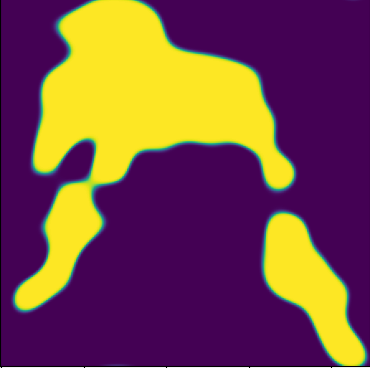}
			\caption{8}
			\label{fig:spectrum_8}
		\end{subfigure}
		\begin{subfigure}[b]{0.25\textwidth}
			\centering
			\includegraphics[width=\textwidth]{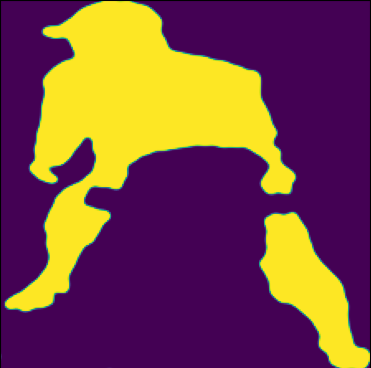}
			\caption{25}
			\label{fig:spectrum_25}
		\end{subfigure}
		
		\caption{ The ground truth vs its reconstructions at various frequencies. }
		\label{fig:spectrums}
	\end{minipage}
\vspace{-0.2cm}	
\end{figure}

To validate that the Fourier Mapping (equation \eqref{eq:FF_output}) works for instance mask prediction, we performed a spectrum analysis on the MS COCO training dataset. Along with verifying our method, this analysis gave us an insight on an optimal number of frequencies for the dataset. We performed this experiment by applying a fast Fourier transform on all the target object masks in the COCO training dataset. This Fourier transform gave us the coefficients of a Fourier series, which hold the same meaning as the coefficients prediction $\textbf{W}$ of FourierMask. Firstly, we sampled only the lower frequency coefficients of the Fourier series and reconstructed the object's mask by applying equation \eqref{eq:FF_output}. We did this for all the objects' masks in COCO training set and evaluated the IoU loss of the reconstruction compared to the target. Then we incrementally added higher frequency coefficients and repeated the above mentioned procedure until we reached the maximum number of frequencies.
The figure \ref{fig:Spectrum} shows the mean IoU loss at various frequencies. It can be seen that the loss decreases exponentially. We choose the maximum frequency as 12 since it has a low enough reconstruction loss and fits comfortably in our GPU memory. Figure \ref{fig:spectrums} illustrates a visual comparison between the ground truth and reconstructions using varying number of frequencies. 
\subsection{Number of frequencies}
\label{ssec:Numfreq}
\begin{figure}[b]
	\vspace{-0.2cm}	
	\centering
	\begin{minipage}[b]{0.42\textwidth}
		\begin{center}
			\includegraphics[width=0.98\linewidth]{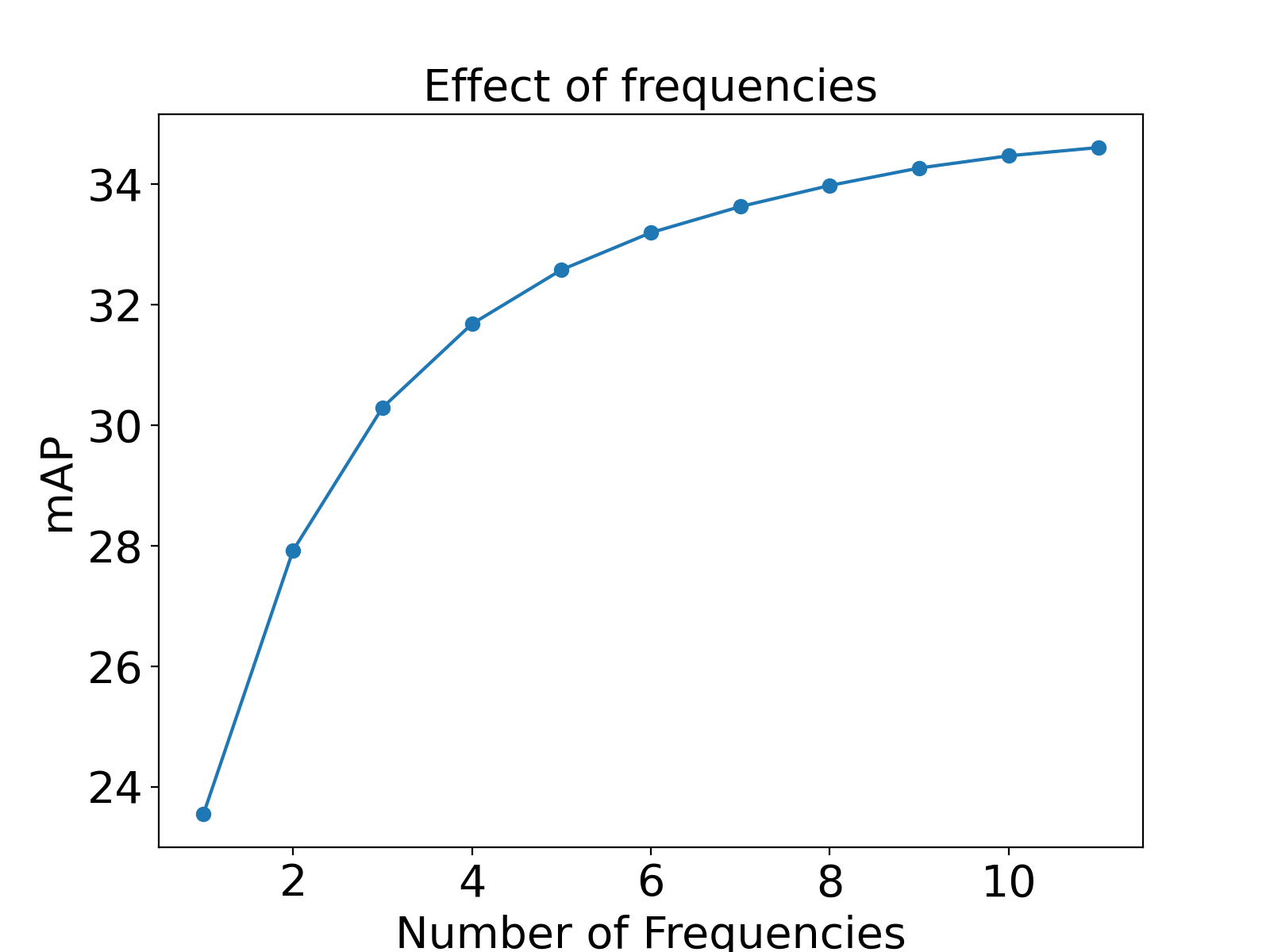}
		\end{center}
		\caption{The mAP when using a subset of trained frequencies.}
		\label{fig:mapvsfreq}	
	\end{minipage}
\hspace{1cm}
	\begin{minipage}[b]{0.4\textwidth}
		\centering
		\begin{subfigure}[b]{0.32\textwidth}
			\centering
			\includegraphics[width=\textwidth]{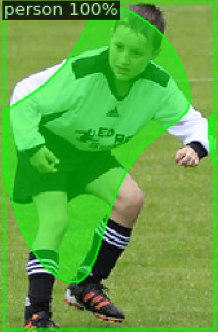}
			\caption{1}
			\label{fig:freq5}
		\end{subfigure}
		\begin{subfigure}[b]{0.32\textwidth}
			\centering
			\includegraphics[width=\textwidth]{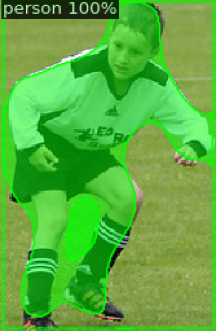}
			\caption{4 }
			\label{fig:freq6}
		\end{subfigure}
		\begin{subfigure}[b]{0.32\textwidth}
			\centering
			\includegraphics[width=\textwidth]{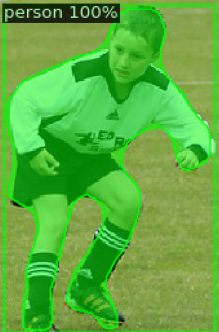}
			\caption{12}
			\label{fig:freq12}
		\end{subfigure}
		\caption{Mask predictions using various frequencies}
		\label{fig:freq}
	\end{minipage}
\vspace{-0.3cm}	
\end{figure}
To validate our experiment from the previous section, we trained a FourierMask with a similar configuration. Rather than prediciting a set of coefficients for each pixel, we modified the architecture to predict a single vector for the whole image. We applied strided (stride=2) 3$\times$3 convolutions 2 times (on the ROI) to reduce the feature size by 1/4th, and then used a fully connected layer to predict the coefficients. The network architecture is shown in the supplementary material. We applied the equation  \eqref{eq:FourierFeatures} and \eqref{eq:FF_output} for generating the mask. We copied the predicted Fourier coefficients $p$ times to match the dimensions for matrix multiplication in equation \eqref{eq:FourierFeatures}. We trained the network with 12 frequencies and an output resolution of $56\times 56$ using IoU loss. In this experiment, we did not add an MLP branch and trained only the equation \eqref{eq:FF_output}. We evaluated the mAP precision of the network on the COCO validation dataset when using a subset of Fourier component frequencies. The network was trained on 12 frequencies, but during inference we incrementally added the higher frequency components starting from the first component. Figure \ref{fig:mapvsfreq} shows the result of this test. The mAP shows a similar trend as seen in figure \ref{fig:Spectrum} and therefore validates the spectrum analysis and the choice of 12 maximum frequencies. The figure \ref{fig:freq} shows an  example how the masks change when different number of frequencies are used.

\subsection{Fourier Features based MLP}
\label{ssec:FFbasedMLP}
\begin{table}[t]
	\vspace{-0.2cm}	
	\begin{center}
		\caption{Comparison of various FourierMask architectures with Mask R-CNN.}
		\label{tab:FFvsMLP}
		\begin{tabular}{|l|c|c|}
			\hline
			Model & Backbone & mAP \\
			\hline\hline
			Mask R-CNN & ResNet-50 & 34.86 \\
			\hdashline
			FF & ResNet-50 & 34.89 \\
			FF + MLP & ResNet-50 & 34.97\\
			FF + MLP & ResNeXt-101 & 39.09 \\
			\hline
		\end{tabular}
	\end{center}
\vspace{-0.3cm}	
\end{table}

To validate that the Fourier Feature based MLP improves the performance of FourierMask, we trained 2 networks with the architecture shown in figure \ref{fig:FourierMaskHead}. In this architecture, the network predicts separate Fourier coefficients for each spatial location. The first network was trained on the masks obtained using $\textbf{y}$ in equation \eqref{eq:FF_output} and output $\textbf{y}^\prime$ of MLP  (FF + MLP). The second network was trained only using equation \eqref{eq:FF_output} (FF). Both networks used 12 Fourier frequencies and had an output resolution of $28\times28$ pixels. We used class agnostic masks and therefore predicted only one class for each region of interest, rather than a mask for each class in the COCO dataset. We had two hidden layers (both with sine activations and 256 neurons) and a single output neuron with sigmoid activation. For the first network (FF + MLP), we took the mean of the masks predicted by $\textbf{y}$ (equation \eqref{eq:FF_output}) and output $\textbf{y}^\prime$ of MLP during inference. The results are shown in the table \ref{tab:FFvsMLP}. As can been seen in the table, the network with an MLP shows the best performance among the models with ResNet-50 backbone. We also trained the same network with a larger ResNeXt-101 \cite{xie2017aggregated} backbone. The improvement of more than 4 mAP over Resnet-50 model shows that our model scales well to bigger backbones.

\subsection{Higher resolution using pixel sub-sampling}
\begin{figure}[t]
	\vspace{-0.2cm}	
	\centering
	\begin{subfigure}[b]{0.18\textwidth}
		\centering
		\includegraphics[width=\textwidth]{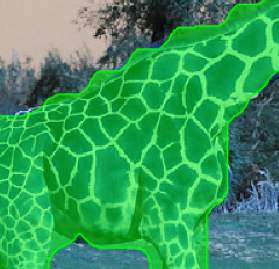}
		\caption{Step = 1}
		\label{fig:freq5}
	\end{subfigure}
	\begin{subfigure}[b]{0.18\textwidth}
		\centering
		\includegraphics[width=\textwidth]{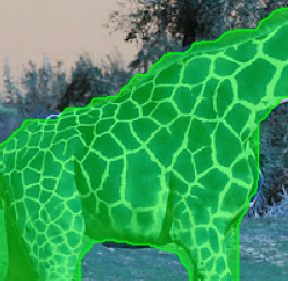}
		\caption{Step = $1/2$}
		\label{fig:freq6}
	\end{subfigure}
	\begin{subfigure}[b]{0.18\textwidth}
		\centering
		\includegraphics[width=\textwidth]{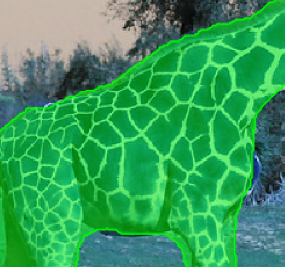}
		\caption{Step = $1/4$}
		\label{fig:freq7}
	\end{subfigure}
	\begin{subfigure}[b]{0.18\textwidth}
		\centering
		\includegraphics[width=\textwidth]{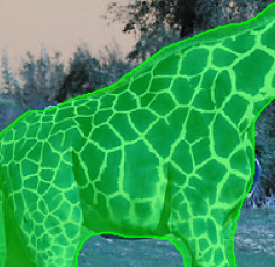}
		\caption{Step = $1/8$}
		\label{fig:freq12}
	\end{subfigure}
	\caption{Subsampling the pixels smooths out the boundaries of the mask.}
	\label{fig:pixstep}
	\vspace{-0.3cm}	
\end{figure}

One of the advantages of our method is that it can predict masks at sub-pixel resolution because implicit representations are continuous in the input domain. We analyzed this by evaluating both the trained networks in section \ref{ssec:FFbasedMLP} on the MS COCO validation set on various pixel steps. For the input $\textbf{x}$ in the equation \eqref{eq:FourierMapping}, rather than using integer values of pixels (pixel step of 1), we used a pixel step of $1/2^{s-1}$, where $s \in \mathbb{Z}^+$ is the scaling factor. This effectively scaled both the height and width of the input $\textbf{x}$ by a factor of $s$. 
To match the size of input pixels $\textbf{x}$, we upsampled the coefficients $\textbf{W}$ in equation \eqref{eq:FourierFeatures} in the spatial dimension using bilinear interpolation, by a scaling factor $2^{s-1}$. Table \ref{tab:maskres} shows the evaluation using the two networks explained in section \ref{ssec:FFbasedMLP}. We can observe that using a lower pixel step improves the mAP. 
\begin{table}[t]
	\vspace{-0.2cm}	
	\begin{center}
		\caption{Sub-sampling performance and speed (GTX 2080Ti).}		
		\label{tab:maskres}
		\begin{tabular}{|l|c|c|c|c|}
			\hline
			Model & Pixel Step & Resolution & mAP & Speed (ms) \\
			\hline\hline
			Mask R-CNN & 1 & 28$\times$28 & 34.86 & 48.7 \\
			\hdashline
			FF & 1 & 28$\times$28 &34.89 & 50.3 \\
			FF & $1/2$ & 56$\times$56 & 35.13 & 59.1 \\
			FF & $1/4$ & 112$\times$112 & \textbf{35.18} & 68.3 \\
			\hdashline
			FF + MLP & 1 & 28$\times$28 & 34.97 & 52.1\\
			FF + MLP & $1/2$ & 56$\times$56 & \textbf{35.18} & 67.0\\
			\hline
			
		\end{tabular}
	\end{center}
\vspace{-0.3cm}		
\end{table}
Figure \ref{fig:pixstep} shows how the mask boundary smooths out when sub sampling the pixels. Note that we trained the network on $28\times 28$ output resolution, but we can generate higher resolution output during inference, which is a considerable advantage over other methods. (see supplementary material for analysis with different number of frequencies at various pixel steps.)
\subsection{Higher resolution using FourierRend}
\label{ssec:subdivision}
To generate higher resolution masks, we used FourierRend (section \ref{ssec:mlp_renderer}) along with the subdivision strategy from Pointrend \cite{Kirillov_2020_CVPR}. We replaced the predictions from equation \eqref{eq:FF_output} (coarse predictions), with the fine detailed predictions from the FourierRend. This resulted in masks which were more crisp and boundary aligned. For training FourierRend, we employ the default settings of point selection strategy along with the point loss from PointRend. The results are shown in the table \ref{tab:subdivision}. Here, we also evaluate the mask quality using the \textit{Boundary IOU} \cite{cheng2021boundary} metric (mAP$_{bound}$), which penalizes the boundary quality more than overall correct pixels. Compared to Mask R-CNN, we see a decent improvement of more than 0.7 mAP$_{mask}$ and 1.6 mAP$_{bound}$ with comparable speeds. We can clearly see visual improvements specially in boundary quality (see figure \ref{fig:vis_compare} and supplementary material). Compared to PointRend, we observe that the masks are more complete  (see figure \ref{fig:vis_compare}) with a reasonably lower inference speed. Note that FourierRend achieves $224\times 224$ in 3 sub-division steps compared to 5 steps of PointRend because FourierRend's initial resolution is $28\times 28$ compared to $7\times7$ of PointRend.
\begin{table}[b]
	\vspace{-0.3cm}	
	\begin{center}
		\caption{The effect of subdivision inference.}		
		\label{tab:subdivision}
		\begin{tabular}{|l|c|c|c|c|c|}
			\hline
			Model & Sub. steps & Resolution & mAP$_{mask}$ &mAP$_{bound}$ & Speed (ms) \\
			\hline\hline
			Mask R-CNN & 0 & 28$\times$28 &34.86 & 21.2 & 48.7 \\
			\hdashline
			FourierRend & 0 & 28$\times$28 &35.01 & 21.0 & 48.7 \\
			FourierRend & 1 & 56$\times$56 & 35.63 & 22.8 & 52.4 \\
			FourierRend & 2 & 112$\times$112& 35.64 & 22.8 & 55.7 \\
			FourierRend & 3 & 224$\times$224 & 35.64 & 22.9 & 59.4 \\
			\hdashline			
			PointRend & 5 & 224$\times$224 & 36.12 & 23.5 & 81.6 \\
			\hline
		\end{tabular}
	\end{center}
\vspace{-0.3cm}		
\end{table}
\section{Conclusion}
In this paper, we show how implicit representations combined with the Fourier series can be applied to the task of instance segmentation.
We illustrate that the masks generated using our Fourier mapping are compact. The lower Fourier frequencies hold the shape and higher frequencies hold the sharp edges. Furthermore, by sub-sampling the pixel coordinates in our implicit MLP, we can generate higher resolution masks during inference, which are visually smoother and improve the mAP over our baseline Mask R-CNN. We also show that our renderer MLP FourierRend improves the boundary quality of FourierMask significantly. See supplementary materials for more details. 

\bibliographystyle{splncs04}
\bibliography{egbib}
\end{document}